\documentclass[lettersize,journal]{IEEEtran}
\usepackage{amsmath,amsfonts}
\usepackage{algorithmic}
\usepackage{algorithm}
\usepackage{array}
\usepackage[caption=false,font=normalsize,labelfont=sf,textfont=sf]{subfig}
\usepackage{textcomp}
\usepackage{stfloats}
\usepackage{url}
\usepackage{verbatim}
\usepackage{graphicx}
\usepackage{cite}

\usepackage[center]{caption}
\usepackage{multirow}
\usepackage{booktabs}
\usepackage{tikz-cd}
\usepackage{hyperref}
\hypersetup{colorlinks,urlcolor=magenta,linkcolor=blue,citecolor=blue}

\begin{document}

\title{DwinFormer: Dual Window Transformers for End-to-End Monocular Depth Estimation}

\author{Md Awsafur Rahman, \IEEEmembership{Student Member, IEEE}, and~Shaikh Anowarul Fattah, \IEEEmembership{Senior~Member, IEEE}

\thanks{M.~A.~Rahman and S.~A.~Fattah are with the Department of Electrical and Electronic Engineering, Bangladesh University of Engineering and Technology, Dhaka-1000, Bangladesh, e-mail: (awsaf49@gmail.com, and fattah@eee.buet.ac.bd).}}

% The paper headers
\markboth{Journal of \LaTeX\ Class Files,~Vol.~XX, No.~XX, Febuary~2023}%
{Shell \MakeLowercase{\textit{et al.}}: A Sample Article Using IEEEtran.cls for IEEE Journals}

\IEEEpubid{This work has been submitted to the IEEE for possible publication. Copyright may be transferred without notice, after which this version may no longer be accessible.}

\maketitle

\begin{abstract} 
Depth estimation from a single image is of paramount importance in the realm of computer vision, with a multitude of applications. Conventional methods suffer from the trade-off between consistency and fine-grained details due to the local-receptive field limiting their practicality. This lack of long-range dependency inherently comes from the convolutional neural network part of the architecture. In this paper, a dual window transformer-based network, namely DwinFormer, is proposed, which utilizes both local and global features for end-to-end monocular depth estimation. The DwinFormer consists of dual window self-attention and cross-attention transformers, Dwin-SAT and Dwin-CAT, respectively. The Dwin-SAT seamlessly extracts intricate, locally aware features while concurrently capturing global context. It harnesses the power of local and global window attention to adeptly capture both short-range and long-range dependencies, obviating the need for complex and computationally expensive operations, such as attention masking or window shifting. Moreover, Dwin-SAT introduces inductive biases which provide desirable properties, such as translational equvariance and less dependence on large-scale data. Furthermore, conventional decoding methods often rely on skip connections which may result in semantic discrepancies and a lack of global context when fusing encoder and decoder features. In contrast, the Dwin-CAT employs both local and global window cross-attention to seamlessly fuse encoder and decoder features with both fine-grained local and contextually aware global information, effectively amending semantic gap. Empirical evidence obtained through extensive experimentation on the NYU-Depth-V2 and KITTI datasets demonstrates the superiority of the proposed method, consistently outperforming existing approaches across both indoor and outdoor environments.
\end{abstract}


\section{Introduction}
\IEEEPARstart{D}{peth} information has a wide variety of applications in different fields~\cite{bokeh, car, robot, ar}, such as refocusing/bokeh, self-driving vehicles, robot motion, augmented reality, and so on. Depth sensors e.g. LiDAR, and ToF are expensive and can capture only low-resolution sparse depth information, which demands the development of single-image (monocular) depth estimation for high-resolution and dense depth information. However, estimating depth relying on a single image is an ill-posed problem~\cite{illpose01} as often there is not enough information to accurately determine the depth of objects in a scene, thus making it very challenging. In the past years, convolutional neural network (CNN) based deep learning methods have dominated monocular depth estimation~\cite{eigen}. But, CNN fails to produce global context-aware pixel-wise prediction due to its intrinsic locality resulting in a trade-off between fine-detailed and consistent depth map~\cite{tradeoff}. Several methods attempted to mitigate these limitations~\cite{deeplab} still the above-mentioned issues persist.

In recent years, transformers~\cite{DPT, TransDepth, Depthformer} have shown promising results for monocular depth estimation. They are notable for their ability to capture long-range dependencies in data. A few years back, Vision Transformer~\cite{ViT} applied pure transformer to the image surpassing the CNN, which verified the usability of transformers in computer vision. Transformers, which have been applied to pixel-wise prediction, have addressed the issue of a local receptive field. Nevertheless, they have also introduced new challenges, including a lack of multi-scale features, the need for higher image resolution than text, and the fact that the computational complexity of self-attention scales quadratically with the size of the image. Later on, several attempts have been made to address these issues with hierarchical transformer~\cite{SwinT} but they provide limited coverage for the global-receptive field (cross-window connection)~\cite{GCViT}. Recent studies such as~\cite{GCViT} and~\cite{MaxViT} made progress towards addressing these issues. Despite that, these studies present their own set of challenges. The utilization of global sparse attention in~\cite{MaxViT} leads to a deterioration in the quality of global features as the size of the image increases. Meanwhile, the study presented in~\cite{GCViT} suffers from ineffective global attention, resulting in a loss of crucial global information and suboptimal performance.

\IEEEpubidadjcol

\begin{figure}[t]
    \centering
    \includegraphics[scale=0.035]{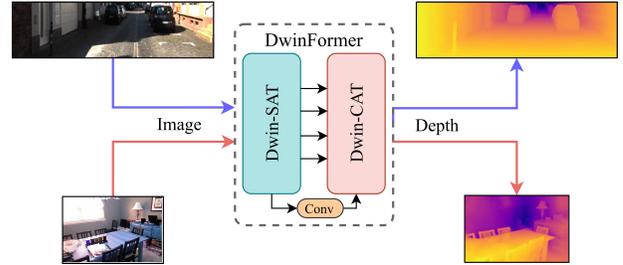}
\captionsetup{justification=centering,singlelinecheck=false}
    \caption{Graphical abstract of the proposed method.}
    \label{fig:graphical-abstract}
    \vspace*{-0.4cm}
\end{figure}

Despite the success of the recent transformer based methods mentioned above in aligning depth edges with object boundaries, they often struggle to accurately assign depth labels to pixels due to problems in effectively fusing encoder and decoder features. Typically a skip connection is used to fuse encoder and decoder features, which applies convolution to features after concatenation. Due to convolution's intrinsic locality, the flow of semantic information is restricted from long ranges affecting the ability of the model to predict the correct depth label for a pixel. To mitigate this issue, skip-attention module is introduced by~\cite{PixelFormer} that integrates encoder-decoder features contextually using local window-based cross-attention. Despite the benefits of transformer-based fusion of encoder-decoder features, the skip-attention module is still restricted by the limited receptive field of local-window attention, thereby it is only able to effectively incorporate information from a limited range of input pixels leaving the global information unused, potentially limiting its overall performance.
In this paper, a transformer based architecture is proposed for end-to-end depth estimation while addressing above mentioned issues of existing approaches. The main contributions of the proposed method can be summarized as follows:
\begin{enumerate}
  \item A dual window transformer-based network, namely DwinFormer, is proposed for end-to-end monocular depth estimation. Here dual window self-attention (Dwin-SAT) and cross-attention (Dwin-SAT) transformers are introduced to effectively capture long-range dependencies and local fine-grained details.
  \item Proposed Dwin-SAT introduces an effective design of a transformer-based backbone, which utilizes both local and global window attentions to mitigate the trade-off between fine details and consistency in depth maps by capturing both local and global contextual information.
  \item To bridge the semantic gap between encoded and decoded features, the proposed Dwin-CAT decoder seamlessly fuses encoder and decoder features with both fine-grained local and contextually aware global information.
\end{enumerate}

\section{Related Work}
\label{sec:related-work}
%\subsection{CNN Encoder and CNN Decoder}
Eigen et al.~\cite{eigen} first proposed a coarse to-fine method to estimate depth. Zhou et al.~\cite{zhou} first introduced a method for predicting the camera's ego-motion and depth map from monocular video. Later on, Laina et al.~\cite{laina} proposed a fully CNN-based residual network, incorporating up-projection block for improved performance. Ranjan et al~\cite{ranjan2019competitive} trained depth estimation in conjunction with optical flow estimation and motion segmentation to achieve synergistic results. Yourun et al.~\cite{zhang2022self} developed a novel multi-scale method that downsamples the predicted depth map and performs image synthesis at multiple resolutions for improved model performance.

In recent years, transformers have been a popular choice for encoders for their ability to capture long-range dependencies. TransDepth~\cite{TransDepth}, and DPT~\cite{DPT} are the first to utilize vision-transformer (ViT) encoder and a convolutional decoder for depth estimation. Later on, SwinDepth~\cite{SwinDepth} utilizes a hierarchical vision-transformer named Swin Transformer~\cite{SwinT} as encoder and multi-scale convolutional  block as decoder. On the other hand, AdaBin~\cite{AdaBin} uses an additional transformer-based block to adaptively estimate depth values for an image by dividing the range of possible depths into bins and calculating the final depth estimates as linear combinations of the bin centers. Very few explorations have been made in this area. But, recently in PixelFormer~\cite{PixelFormer}, a window attention-based transformer is utilized to effectively fuse the features produced by encoder and decoder. This approach avoids the limitations of simply concatenating these features, which can incite semantic gap~\cite{zioulis2022hybrid}. Despite using a transformer-based fusion of encoder-decoder features, this method fails to reach its true potential as its skip-attention module is highly impeded by incorporation of information from a limited range of input pixels due to local-window attention. 

In essence, CNNs struggle with a trade-off between fine details and consistency in depth maps, while transformers have shown promise in capturing global information, yet they are hindered by high computational complexity, lack of multi-scale features, and inductive bias. Recent attempts to overcome these limitations through different types of window attention face new sets of challenges, such as limited global coverage, impaired global context, and loss of global context. Meanwhile, the fusion of encoder and decoder features remains a challenge due to semantic-gap, and while skip-attention module~\cite{PixelFormer} has improved the fusion process, it is still limited by its local receptive fields.

\section{Methodology}
\label{sec:methodology}
\begin{figure*}[t]
    \centering
\captionsetup{justification=justified,singlelinecheck=false}
    \includegraphics[scale=0.054]{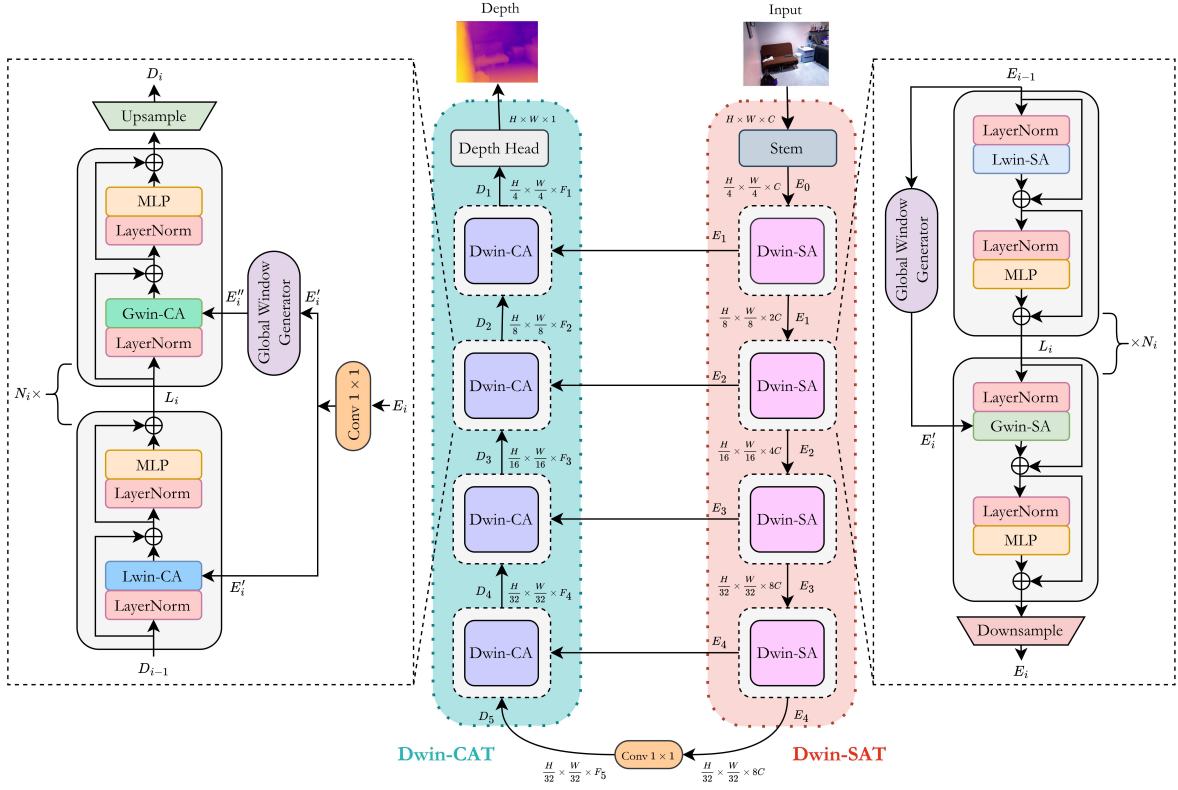}
    \caption{Graphical overview of the DwinFormer Architecture. The input image is encoded into multiscale feature maps using a series of Dwin-SA blocks in the Dwin-SAT. The encoded features are then reconstructed into a depth map by the Dwin-CAT, which integrates features from Dwin-SA of the same level and Dwin-CA of the previous level. The final depth map is produced by the Depth Head, processing the decoded features.}
    \label{fig:arch}
\end{figure*}
\subsection{Problem Definition}
The proposed method models the depth map of an RGB image as the probability of maximum depth for all pixels. Given an image $I \in \mathbb{R}^{H \times W \times 3}$, a model $\mathcal{F}(\Theta)$ is developed to predict a probability mask $\hat{p} \in [0, 1]^{H \times W \times 1}$ indicating the likelihood of the maximum depth for all pixels in the image. The mask is transformed into the depth map $\hat{y} = \hat{p} \odot \textit{max\_depth}$. The model's parameters $\Theta$ are optimized by minimizing a chosen loss function $\mathcal{L}(\Theta, y, \hat{y})$ with respect to the true depth map $y$ through $\Theta^* = \arg\min_{\Theta} \mathcal{L}(\Theta, y, \hat{y})$.

\subsection{Overview of Proposed DwinFormer Architecture}
The proposed architecture, as shown in Fig.~\ref{fig:arch}, introduces the Dual Window Self Attention Transformer (Dwin-SAT) backbone to process the input image $I$. The Dwin-SAT backbone employs multiple layers of Dual Window Self Attention (Dwin-SA) to extract feature maps representing the image at different resolution scales. These resolution scales are represented by $i$ and are defined as $\frac{1}{2^{i + 1 }}$ relative to the input image $I$, where $i \in {\{1, 2, 3, 4\}}$ and $i=5$ has same resolution as $i=4$. Dwin-SA leverages both Local Window Self Attention (Lwin-SA) and Global Window Self Attention (Gwin-SA), to extract both locally and globally contextualized features from the input image. These features are subsequently refined through the proposed Dual Window Cross Attention (Dwin-CA), which is employed at different stages to merge the encoder-decoder feature maps hierarchically. Dwin-CA exploits Local Window Cross Attention (Lwin-CA) and Global Window Cross Attention (Gwin-CA) respectively to fuse encoder-decoder features with both local and global contexts. Finally, the resulting decoded features from the stack of Dwin-CA layers are processed by the Depth Head module to estimate the per-pixel probability of maximum depth. In what follows, the blocks of DwinFormer are explained.

\subsection{Proposed Encoder Architecture: Dwin-SAT}
\begin{figure}[t]
    \centering
    \includegraphics[scale=0.027]{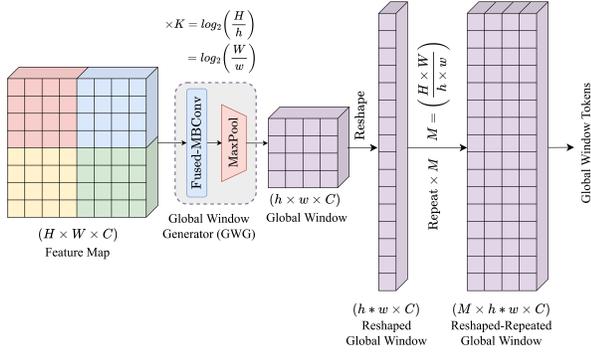}
\captionsetup{justification=justified,singlelinecheck=false}
    \caption{The input feature map with $H, W$ dimensions, is fed into the Global Window Generator (GWG), which uses the Fused-MBConv and MaxPool layer repeatedly $K$ times to generate a global window with $h, w$ dimensions which is reshaped then repeated $M$ times.}
    \label{fig:gwg}
\end{figure}

The proposed Dwin-SAT backbone, as depicted in Fig.~\ref{fig:arch}, effectively extracts features at multiple resolutions by iteratively downgrading the spatial dimensions and upgrading the channel dimensions by factors of 2. The process begins with the Stem block, which generates overlapping patches from the input image $I$ to facilitate the application of the transformer. These patches are then projected into a $C$-dimensional embedding space through the use of a $3\times3$ convolutional layer with a stride of $2\times2$ and $1\times1$ padding. Subsequently, the number of patches is reduced by a factor of 2 using a strided convolution, thus allowing the Dwin-SAT backbone to continuously extract features at different levels of detail. Finally, the resultant features from Dwin-SAT are operated by a $1 \times 1$ convolution to precisely control the channel of the decoder.
Specifically, Dwin-SAT utilizes the Dwin-SA module to extract spatial characteristics by interleaving between Lwin-SA and Gwin-SA modules with local and global contexts respectively. In contrast to Gwin-SA, which operates global attention using both global and local windows, Lwin-SA operates local attention using only local windows in the manner described in Swin Transformer~\cite{SwinT}. The global window in Gwin-SA is generated by Global Window Generator (GWG) from the whole input feature maps and shares across all local windows, thus resulting in a reduced number of parameters and FLOPs. It confines information across the entire input feature maps for interaction with local query features. Specifically, as shown in Fig.~\ref{fig:gwg}, the generator consists series of a Fused-MBConv~\cite{GCViT} block followed by a max pooling layer. Then, using a Downsample block at the end of each stage, the spatial dimension of resulting features is decreased and the channel dimension is increased by a factor of $2$. The Downsample block follows the same architecture in~\cite{GCViT} comprising a modified Fused-MBConv block, followed by a $2\times2$ strided convolution with a kernel size of $3\times3$ and layer normalization layer, which employs spatial feature contraction from CNN that impose inductive bias and inter-channel interactions. At the end of the Dwin-SAT, a $1 \times 1$ convolution is used to control the number of output channels of each decoder layer at each scale. For a given scale $i$, the encoder can be expressed as

\begin{equation}
\begin{aligned}
\label{eq:mwin-sat}
{}_{1}Q^e_l & = LayerNorm(E_{i-1}) \\
K^e_l & = MLP(E_{i-1}) ; V^e_l = MLP(E_{i-1}) \\
L_i & = \operatorname{Lwin-SA}({}_{1}Q^e_l, K^e_l, V^e_l) + E_{i-1} \\
L_i & = L_i + MLP(LayerNorm(L_i)) \\
{}_{2}Q^e_l & = LayerNorm(L_i) ; E'_i = GWG(E_{i-1}) \\
K^e_g & = MLP(E'_i) ; V^e_g = MLP(E'_i) \\
E_i & = \operatorname{Gwin-SA}({}_{2}Q^e_l, K^e_g, V^e_g) + L_i \\
E_i & = E_i + MLP(LayerNorm(E_i))\\
E_i & = \operatorname{Downsample(E_i)}\\
\end{aligned}
\end{equation}

\subsection{Dual Window Self Attention (Dwin-SA)}
\label{subsection-mwinsa}
\begin{figure*}[t]
    \centering
    \includegraphics[scale=0.055]{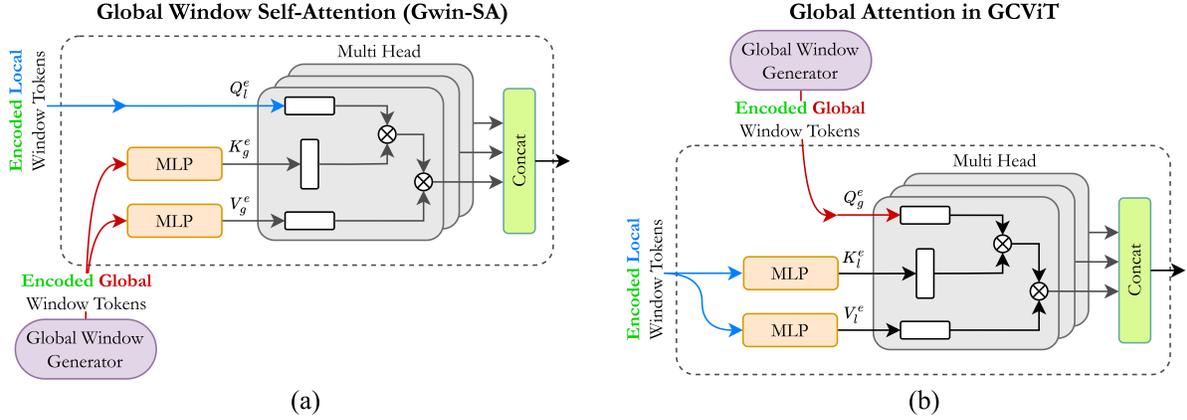}
    \captionsetup{justification=justified,singlelinecheck=false}
    \caption{Comparison between Global Window Self Attention in proposed Dwin-SAT and Global Attention in GCViT. While both methods use attention on encoded window tokens, Gwin-SA leverages global key-values with local queries and GCViT uses global queries with local key-values. Here, Global Window Generator is used to create global query-key-values.}
     \label{fig:sa}
\end{figure*}

The Dwin-SAT backbone's primary computational operator is Dwin-SA. In general, it operates through the process of contextualization, which involves computing the pairwise relations between each query position and all key positions to form an attention map. This map serves as a weighting function, which is used to linearly combine the values to produce the output at each query position. In short, the output at each query position is expressed as a weighted sum of the values, where the weights are computed using the query and key through the attention mechanism. The Dwin-SA block employs two sub-mechanisms for feature extraction. Firstly, Lwin-SA mechanism partitions the image into smaller windows (referred to as local windows) and performs self-attention between patches within these windows to extract fine-grained local features. Secondly, the Gwin-SA, a novel attention mechanism, exploits the global window to interact with patches located anywhere in the image, allowing it to consider information outside of the local window. Gwin-SA accesses global information from the image through global keys and values generated from GWG, which represents the entire image. This enables Gwin-SA to take into account the global context when making attention decisions. Fig.~\ref{fig:sa} provides a visual insight into these attentions and they also can be expressed as follows,

\begin{align}
\label{eq:lwin-sa}
& \operatorname{Lwin-SA}\left(Q^e_l, K^e_l, V^e_l\right)=S\left(Q^e_l {K^e_l}^T/\sqrt{d}+B\right)V^e_l
\end{align}
\begin{align}
\label{eq:gwin-sa}
& \operatorname{Gwin-SA}\left(Q^e_l, K^e_g, V^e_g\right)=S\left(Q^e_l {K^e_g}^T/\sqrt{d}+B\right)V^e_g
\end{align}

where $Q^e_l, K^e_l, V^e_l$ denotes to query, key, and value in the local window from encoder whereas $K^e_g, V^e_g$ denotes to key and value in the global window from encoder generated by GWG. For both local and global windows in encoder, $Q, K, V \in \mathbb{R}^{M^2 \times d}$; $d$ is the query/key dimension, $M^2$ is the number of patches in a window, and $S$ is the Softmax function. Following~\cite{SwinT, GCViT}, assuming relative position along each axis lies in the range $[-M+1, M-1]$, learnable relative position bias $B$ is sampled from bias matrix $\hat{B} \in \mathbb{R}^{(2 M-1) \times(2 M-1)}$.

\begin{figure*}[t]
    \centering
    \includegraphics[scale=0.055]{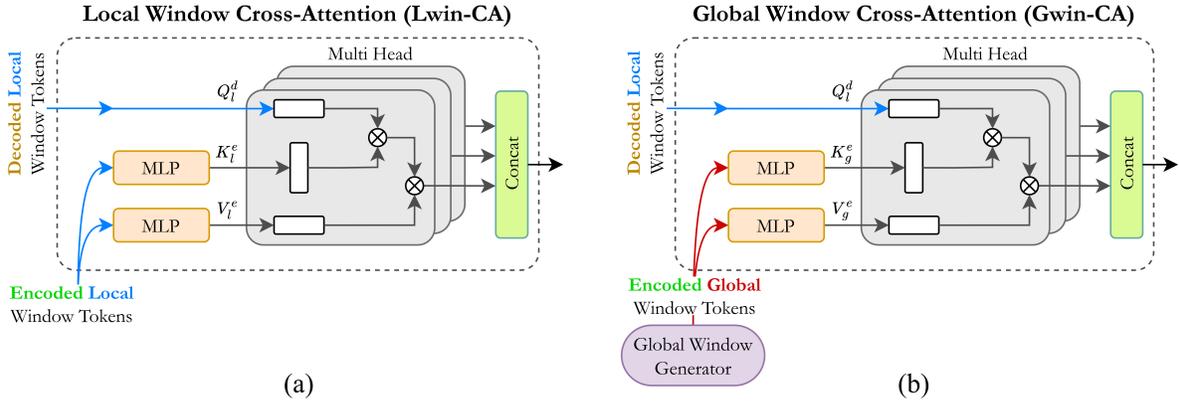}
    \captionsetup{justification=justified,singlelinecheck=false}
    \caption{The Local and Global Window Cross Attention (Lwin-CA and Gwin-CA) blocks in the proposed Dwin-CAT utilize cross-attention between the encoded and decoded features. Both blocks employ decoded features as local queries and encode features as key-values. However, Lwin-CA uses encoded features as local key-values, while Gwin-CA employs them as global key-values via Global Window Generator.}
    \label{fig:ca}
    \vspace*{-0.4cm}
\end{figure*}

The proposed Gwin-SA mechanism draws inspiration from GCViT~\cite{GCViT} but utilizes a distinctive approach to calculate attention as shown in Fig.~\ref{fig:sa}. The global attention in GCViT utilizes global query and a local key-value to contextualize the global window tokens with respect to the local window tokens. Even so, this process can lead to a loss of non-locality for the global window tokens, as the global query only contributes to the weights of the attention and the local value provides the image features. In this approach, the output at each query position is computed as a weighted average of the values, with the weights derived from the query and key through an attention mechanism. In contrast, the Gwin-SA mechanism employs a reverse approach by using a local query and a global key-value to contextualize the local window tokens with respect to the global window tokens. This approach enables the local window tokens to acquire non-locality in attention through the contribution of the global value's supply of image features. A further advantage of the Gwin-SA mechanism lies in its ability to consider each smaller zone by giving deeper effort, and maximize global information, the Gwin-SA mechanism provides a more in-depth effort. The key differences between the GCViT approach and the Gwin-SA mechanism are summarized as follows:

\begin{itemize}
\item Global Attention in GCViT:
\begin{itemize}
\item[$\Rightarrow$] Global query + local key-value
\item[$\Rightarrow$] Contextualize global window w.r.t local window
\end{itemize}

\item Gwin-SA in DwinFormer:
\begin{itemize}
\item[$\Rightarrow$] Local query + global key-value
\item[$\Rightarrow$] Contextualize local window w.r.t global window
\end{itemize}
\end{itemize}

\subsection{Proposed Decoder Architecture: Dwin-CAT}

The proposed Dual Window Cross Attention Transformer (Dwin-CAT) is depicted in Fig.~\ref{fig:arch}. Here, the number of output channels of each decoder layer at each resolution scale $i$ is represented as $F_{i} = 2^{i+4}$, where $i \in {1, 2, 3, 4, 5}$. The decoder takes the lowest-resolution feature map from the encoder and gradually enhances it to produce a final depth representation. At each level, Dwin-CAT integrates details from the encoder's feature map through skip connections and upscaling. However, the encoded feature with its rich information can sometimes mismatch with the decoded feature that holds more semantic information, making the optimization process challenging. To bridge this semantic gap, Dwin-CAT uses the Dwin-CA component that integrates both local and global contexts, resulting in a more contextualized feature. The encoded features are first passed through a $1\times1$ convolution to harmonize their channels with the previous decoded features and facilitate attention. The Dwin-CA component in Dwin-CAT fuses the encoded and decoded features using Local Window Cross Attention (Lwin-CA) and Global Window Cross Attention (Gwin-CA). The resulting feature is passed through the Upsample layer to simultaneously upscale the spatial dimension and downscale the channel dimension. The decoder operates similarly to Dwin-SAT, but it uses cross-attention between the encoded and decoded features, where queries come from the encoder and keys-values from the decoder. In mathematical terms, the decoder layer can be expressed as follows,

\begin{equation}
\begin{aligned}
\label{eq:mwin-cat}
% D_i & = f^{d}_{i}(D_{i-1}, E_i) \\
{}_{1}Q^d_l & = LayerNorm(D_{i-1}); E'_i = Conv_{1 \times 1}(E_i) \\
K^e_l & = MLP(E'_i) ; V^e_l = MLP(E'_i) \\
L_i & = \operatorname{Lwin-CA}({}_{1}Q^d_l, K^e_l, V^e_l) + D_{i-1} \\
L_i & = L_i + MLP(LayerNorm(L_i)) \\
{}_{2}Q^d_l & = LayerNorm(L_i) ; E''_i = GWG(E'_i) \\
K^e_g & = MLP(E''_i) ; V^e_g = MLP(E''_i) \\
D_i & = \operatorname{Gwin-CA}({}_{2}Q^d_l, K^e_g, V^e_g) + L_i \\
D_i & = D_i + MLP(LayerNorm(D_i))\\
D_i & = \operatorname{Upsample(D_i)}\\
\end{aligned}
\end{equation}

\subsection{Dual Window Cross Attention (Dwin-CA)}

The Dual Window Cross Attention (Dwin-CA) is an effective technique that seamlessly integrates encoder and decoder features while addressing semantic discrepancies between them. Similar to Dual Window Self-Attention (Dwin-SA), it utilizes local and global window attention mechanisms. However, instead of self-attention within encoded features, it utilizes cross-attention between encoded and decoder features. This allows Dwin-CA to bridge semantic gaps between the rich and nuanced encoded feature map and the decoded feature map, which holds the prevalence of information about the final depth representation. The technique does this by incorporating both local and global context through Local Window Cross Attention (Lwin-CA) and Global Window Cross Attention (Gwin-CA) respectively. Lwin-CA extracts fine-grained local features by performing cross-attention between the encoder and decoder using local windows while Gwin-CA captures long-range dependencies by performing cross-attention between the encoder and decoder using global windows. Fig.~\ref{fig:ca} shows an illustration of the proposed cross-attention mechanism and they can also be mathematically represented as
\begin{align}
\label{eq:lwin-ca}
& \operatorname{Lwin-CA}\left(Q^d_l, K^e_l, V^e_l\right)=S\left(Q^d_l {K^e_l}^T/\sqrt{d}+B\right)V^e_l
\end{align}
\begin{align}
\label{eq:gwin-ca}
& \operatorname{Gwin-CA}\left(Q^d_l, K^e_g, V^e_g\right)=S\left(Q^d_l {K^e_g}^T/\sqrt{d}+B\right)V^e_g
\end{align}
The notations used in these equations, such as $Q^d_l$, $K^e_l$, $V^e_l$, $K^e_g$, $V^e_g$, $d$, $B$, and $S$ are consistent with the previously defined notations in equation \ref{eq:lwin-sa} and \ref{eq:gwin-sa}, with the exception of $Q^d_l$, which represents the query in the local window from the decoder.

\subsection{Depth Head}
The Depth Head of the proposed architecture utilizes a depth feature map generated by the Dwin-CAT of size $(\frac{H}{4}, \frac{W}{4}, C)$. The depth feature map is then passed through a $3 \times 3$ convolutional layer which transforms it into a probability map of maximum depth for each pixel. Then the depth map is obtained by applying the sigmoid function on this probability map, resulting in a continuous estimate of the depth at each pixel. Finally, the resultant depth map is upsampled $4 \times$ to match the resolution of the ground truth.

\subsection{Loss Function}
The Scale-Invariant loss (SI)~\cite{eigen} is utilized in a scaled form in this paper. The mathematical representation of the loss is given by the equation:

\begin{equation}
\begin{aligned}
\label{eq:si-loss}
\mathcal{L}_{\text {pixel }}=\alpha \sqrt{\frac{1}{T} \sum_{i} g_i^2-\frac{\lambda}{T^2}\left(\sum_i g_i\right)^2}
\end{aligned}
\end{equation}

Where $d_i$ is the ground truth depth, $\hat{d}_i$  is the estimated depth, $T$ represents the number of pixels with valid ground truth values and $g_i$ is computed as $g_i=\log_e(\hat{d}_i)-\log_e(d_i)$. The parameters $\lambda$ and $\alpha$ used in the experiments are set to 0.85 and 10 respectively.

\begin{table}
\centering
\captionsetup{justification=raggedright,singlelinecheck=false}
\caption{Result on NYUv2 Data. The best result is indicated in \textbf{bold}, second best is \underline{underlined}, and symbols $\uparrow$ or $\downarrow$ denote higher/lower values are preferable}
\label{tab:nyuv2-method}
\begin{tabular}{lcccc} 
\toprule
Method                & Abs Rel $\downarrow$ & RMS $\downarrow$ & $\log_{10}$ $\downarrow$ & $\delta_1 \uparrow$  \\ 
\midrule
Eigen et al.~\cite{eigen}         & 0.158                & 0.641            & -                      & 0.769                \\
DORN~\cite{DORN}                 & 0.115                & 0.509            & 0.051                  & 0.828                \\
Yin et al.~\cite{Yin}           & 0.108                & 0.416            & 0.048                  & 0.872                \\
BTS~\cite{BTS}                  & 0.110                & 0.392            & 0.047                  & 0.885                \\
TransDepth~\cite{TransDepth}           & 0.106                & 0.365            & 0.045                  & 0.900                \\
DPT~\cite{DPT}                  & 0.110                & 0.367            & 0.045                  & 0.904                \\
Adabins~\cite{AdaBin}              & 0.103                & 0.364            & 0.044                  & 0.903                \\
P3Depth~\cite{P3depth}              & 0.104                & 0.356            & 0.043                  & 0.898                \\
SwinDepth~\cite{SwinDepth}            & 0.100                & 0.354            & 0.042                  & 0.909                \\
Depthformer~\cite{Depthformer}          & 0.100                & 0.345            & -                      & 0.911                \\
NeWCRFs~\cite{NewCRFs}             & 0.095                & 0.334            & 0.041                  & 0.922                \\
PixelFormer~\cite{PixelFormer}          & \underline{0.090}        & \underline{0.322}    & \underline{0.039}          & \underline{0.929}        \\ 
\midrule
DwinFormer (Proposed) & $\mathbf{0.081}$     & $\mathbf{0.280}$ & $\mathbf{0.034}$       & $\mathbf{0.951}$     \\
\bottomrule
\end{tabular}
\end{table}

\begin{table}
\centering
\captionsetup{justification=justified,singlelinecheck=false}
\caption{Result on KITTI Data. The best result is indicated in \textbf{bold}, second best is \underline{underlined}, and symbols $\uparrow$ or $\downarrow$ denote higher/lower values are preferable}
\label{tab:kitti-method}
\begin{tabular}{lccccccc} 
\toprule
Method           & Abs Rel $\downarrow$ & RMS $\downarrow$   & $\log_{10} \downarrow$ & $\delta_1 \uparrow$  \\ 
\midrule
Eigen et al.~\cite{eigen}               & $0.203$              & $6.307$             & $0.282$              & $0.702$                           \\
DORN.~\cite{DORN}                          & $0.072$              & $2.727$             & $0.120$              & $0.932$                           \\
Yin et al.~\cite{Yin}                          & $0.072$              & $3.258$             & $0.117$              & $0.938$             \\
BTS~\cite{BTS}                             & $0.059$              & $2.756$             & $0.096$              & $0.956$           \\
TransDepth~\cite{TransDepth}                      & $0.064$              & $2.755$             & $0.098$              & $0.956$             \\
Adabins~\cite{AdaBin}                        & $0.058$              & $2.360$             & $0.088$              & $0.964$             \\
DPT~\cite{DPT}                               & $0.060$              & $2.573$             & $0.092$              & $0.959$             \\
SwinDepth~\cite{SwinDepth}                       & 0.064                & 2.643               & -                    & 0.957               \\
Depthformer~\cite{Depthformer}                     & 0.058                & 2.285               & -                    & 0.967               \\
NeWCRFs~\cite{NewCRFs}                         & $0.052$              & $2.129$            & $0.079$              & $0.974$             \\
PixelFormer~\cite{PixelFormer}                     & $\underline{0.051}$  & $\underline{2.081}$ & $\underline{0.077}$  & $\underline{0.976}$ \\ 
\midrule
DwinFormer (Proposed)   & $\mathbf{0.047}$     & $\mathbf{1.959}$    & $\mathbf{0.073}$     & $\mathbf{0.980}$  \\
\bottomrule
\end{tabular}
\end{table}

\section{Experiments}
\label{sec:experiments}

Numerous experiments using the KITTI~\cite{kitti} and NYU Depth V2~\cite{nyuv2} datasets are conducted to confirm the effectiveness of the novel method proposed. Through a thorough analysis of both quantitative and qualitative criteria, the proposed method is put through a thorough evaluation and comparison against current state-of-the-art techniques. Additionally, an ablation study is carried out to clearly show the importance of each individual component. 

\subsection{Datasets}
NYU Depth V2 is an indoor dataset of 120K RGB-depth pairs from 464 scenes, with an official training/testing split of 50K/654 images, and a depth upper bound of 10 meters. The proposed method is trained with a resolution of $480 \times 640$.

KITTI is an outdoor dataset of stereo images from 61 scenes, with a training/testing split of 26K/697 images defined by~\cite{eigen}, and a depth upper bound of 80 meters. The proposed method is trained with a resolution of $704 \times 352$.

\subsection{Implementation Details}
The present study employs the TensorFlow framework to implement its proposed method. Adam optimizer~\cite{Adam}, with a batch size of 8 is utilized during training. Both the KITTI and NYUV2 datasets are trained for a period of 30 epochs, with an initial learning rate of $3 \times 10^{-6}$, that is gradually increased linearly to $2.5 \times 10^{-5}$ and subsequently decreased with a Cosine schedule. The proposed method utilizes 8 x NVIDIA V100 GPUs for training. To prevent overfitting, various data augmentation techniques such as random rotation, horizontal flipping, brightness-contrast-hue adjustments, grayscale, and noise are applied. The proposed architectural design allows for the utilization of pre-trained weights from GCViT for initializing the encoder backbone, thus avoiding the need for training the backbone on ImageNet from scratch. The number of output channels in each level for both the encoder and decoder is dependent on the size of the architecture, where for the XXTiny, XTiny, Tiny, Small, Base, and Large sizes, $C$ is set to $64$, $64$, $64$, $96$, $128$ and $192$ respectively. The remaining architectural parameters are kept similar to those of GCViT~\cite{GCViT}.

\subsection{Evaluation Metrics}
To evaluate and compare the performance of the proposed method in comparison to existing methods five metrics~\cite{eigen} are employed, namely Average Relative error (Abs Rel): $\frac{1}{n} \sum_{i \in n} \frac{\left|d_i-\hat{d}_i\right|}{\hat{d}_i}$; Root Mean Squared error (RMS): $\sqrt{\frac{1}{n} \sum_{i \in n}\left\|d_i-\hat{d}_i\right\|^2}$; Average $\left(\log _{10}\right)$ error: $\frac{1}{n} \sum_{i \in n} \left|\log _{10}\left(d_i\right)-\log _{10}\left(\hat{d}_i\right)\right| ;$ Threshold Accuracy $\left(\delta_i\right)$ : $\%$ of $d_i$ s.t. $\quad \max \left(\frac{d_i}{\hat{d}_i}, \frac{\hat{d}_i}{d_i}\right)=\delta<t h r$ for $thr=$ $1.25$; Squared Relative difference (Sq Rel): $\frac{1}{n} \sum_{i \in n} \frac{\left\|d_i-\hat{d}_i\right\|^2}{\hat{d}_i}$. Here $n$ is the total number of pixels for each depth map, $d_i$, and $\hat{d}_i$ denotes ground truth, and predicted depth for $i$-th pixel respectively.

\subsection{Performance Comparison with Existing Methods}

\subsubsection{Quantitative Analysis}
Tables~\ref{tab:nyuv2-method}, and~\ref{tab:kitti-method} present a quantitative analysis, respectively on the outdoor dataset KITTI and indoor dataset NYUv2, comparing the proposed method to existing techniques. The results of the comparison reveal that the proposed method exhibits outstanding performance, surpassing the other methods by a considerable margin across a wide range of metrics. The only exception is the metric $\delta_3$ where the results are highly saturated, but even in that case, the proposed method's results are competitive. These results conclusively demonstrate the exceptional performance of the proposed method in the field of monocular depth estimation, and its ability to produce highly accurate and precise depth maps.

\subsubsection{Qualitative Analysis}
\begin{figure*}[t]
    \centering
    \captionsetup{justification=centering}
    \includegraphics[scale=0.17]{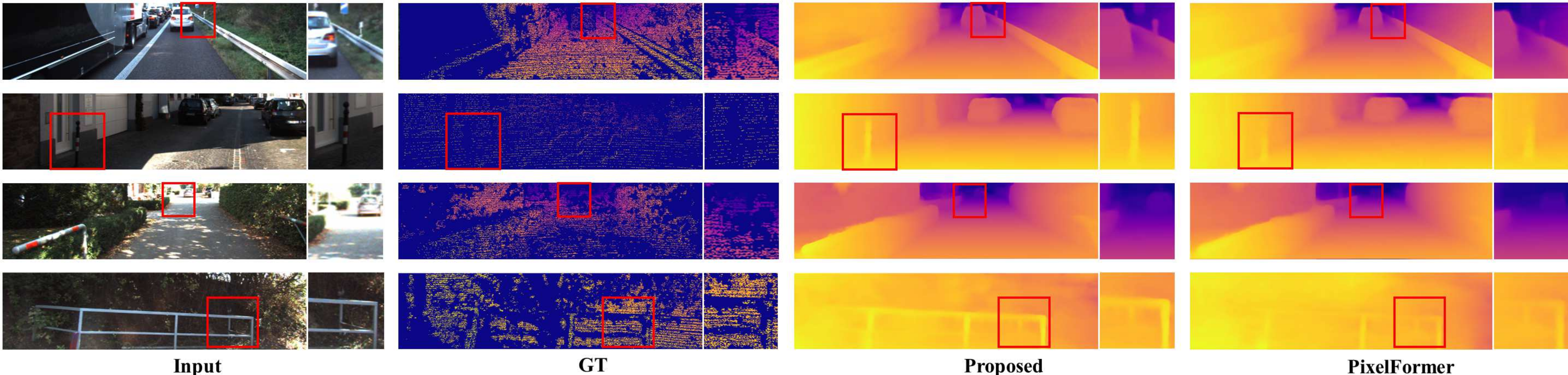}
    \caption{Qualitative comparison between previous SOTA and proposed method for KITTI dataset.}
    \label{fig:kitti-quality}
\end{figure*}

\begin{figure}[h]
    \centering
    \captionsetup{justification=raggedright}
    \includegraphics[scale=0.085]{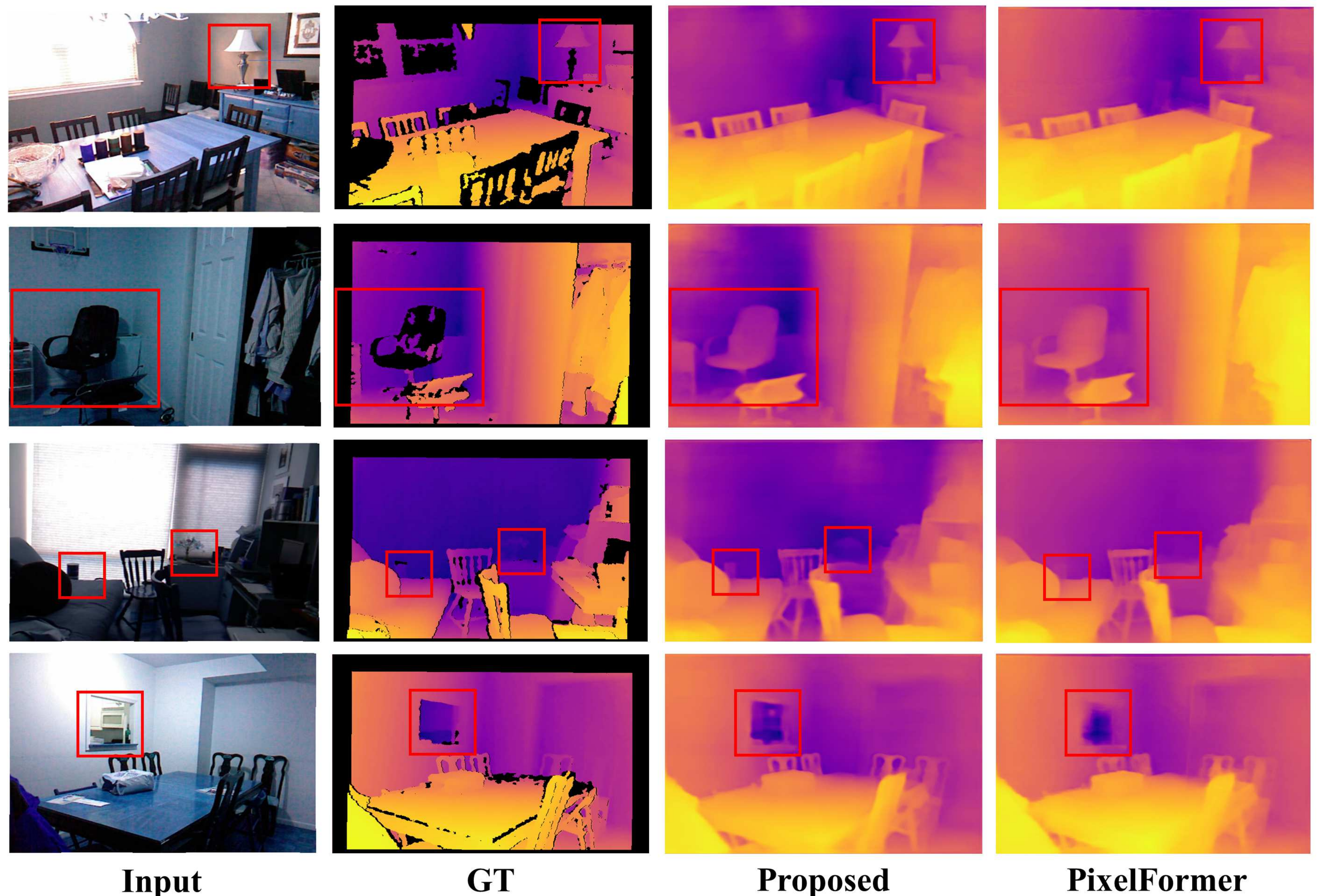}
    \caption{Qualitative comparison between previous SOTA and proposed method for NYUv2 dataset.}
    \label{fig:nyuv2-quality}
\end{figure}

Figures~\ref{fig:kitti-quality} and \ref{fig:nyuv2-quality} present an insightful comparison between the proposed method and existing techniques, respectively, on the outdoor dataset KITTI and indoor dataset NYUv2. As it is plainly evident from the figures, the proposed method demonstrates an aptitude for effectively identifying depth cues such as object boundaries, and sharp edges while producing both consistent and detailed depth maps even for objects with missing information in the RGB image. This highlights the superiority of the proposed method.

\subsection{Ablation Study}

\subsubsection{Effectiveness of Dwin-SAT}

Table~\ref{tab:kitti-mwin-sa} presents a comprehensive examination of the performance of the proposed Dwin-SAT in comparison to other established backbones. In this ablation study, Dwin-CAT is utilized as decoders across all encoders for comparison purposes.  It is manifestly evident that the proposed Dwin-SAT outperforms the other methods significantly. This is attributed to its unique ability to effectively extract features with both local and global contexts, thus conclusively validating the efficacy of the proposed architecture.

\begin{table}
\centering
\captionsetup{justification=raggedright,singlelinecheck=false}
\caption{Ablation study of Dwin-SAT on KITTI Eigen Split. Results are denoted by $\uparrow$ for better and $\downarrow$ for worse. The best results are in \textbf{bold}}
\label{tab:kitti-mwin-sa}
\scalebox{1.1}{
\begin{tabular}{lccc} 
\toprule
Method                    & Abs Rel $\downarrow$ & Sq Rel $\downarrow$ & $\delta_1 \uparrow$  \\ 
\midrule
EfficientNet-B7~\cite{EfficientNet}           & $0.056$              & $0.165$             & $0.968$              \\
SwinT-Large~\cite{SwinT}             & $0.053$              & $0.153$             & $0.972$              \\
ConvNeXt-Large~\cite{ConvNeXt}            & $0.052$              & $0.154$             & $0.974$              \\
MaxViT-Large~\cite{MaxViT}              & $0.051$              & $0.140$             & $0.975$              \\
GCViT-Large~\cite{GCViT}               & $0.050$              & $0.138$             & $0.977$              \\ 
\midrule
Dwin-SAT (Proposed) & $\mathbf{0.047}$     & $\mathbf{0.136}$    & $\mathbf{0.980}$     \\
\bottomrule
\end{tabular}
}
\end{table}

\subsubsection{Effectiveness of Dwin-CAT}

Table~\ref{tab:kitti-mwin-ca} presents the performance of our proposed Dwin-CAT module in comparison to existing techniques. In this ablation study, Dwin-SAT is utilized as encoders across all decoders for comparison purposes. The results clearly demonstrate that the proposed Dwin-CAT module stands out as a clear winner, exhibiting a significant superiority over the other methods, thus validating the efficacy of the proposed module.

\begin{table}
\centering
\captionsetup{justification=raggedright,singlelinecheck=false}
\caption{Ablation study of Dwin-CAT on KITTI Eigen Split. Results are denoted by $\uparrow$ for better and $\downarrow$ for worse. The best results are in bold.}
\label{tab:kitti-mwin-ca}
\scalebox{1.19}{
\begin{tabular}{lccc} 
\toprule
Method              & Abs Rel $\downarrow$         & Sq Rel $\downarrow$         & $\delta_1 \uparrow$          \\ 
\midrule
Add-Conv            & $0.060$                     & $0.176$                     & $0.961$                      \\
Concat-Conv         & $0.058$                     & $0.174$                     & $0.963$                      \\
Nested-Skip~\cite{Unet++}         & $0.056$ & $0.168$ & $0.965$  \\
SAM~\cite{PixelFormer}   & $0.052$                     & $0.160$                     & $0.969$                      \\ 
\midrule
Dwin-CAT (Proposed) & $\mathbf{0.047}$             & $\mathbf{0.136}$            & $\mathbf{0.980}$             \\
\bottomrule
\end{tabular}
}
\end{table}

\section{Conclusion}
\label{sec:conclusion}
In this paper, a transformer-based design namely, DwinFormer is introduced for end-to-end monocular depth estimation. The proposed encoder and decoder incorporate both self-attention and cross-attention mechanisms, utilizing both local and global receptive fields. This approach effectively addresses challenges, such as trade-off between consistency and fine details in the depth map and bridging semantic disparities between the encoded and decoded features. As a result, efficient feature extraction and improved reconstruction of depth maps are acquired, leading to more accurate depth maps. The experimental results firmly establish the dominance of the proposed method over existing approaches, as seen by the remarkable performance improvement on the NYUv2 and KITTI datasets.

\bibliographystyle{IEEEtran}
\bibliography{REFERENCE}

\end{document}